\documentclass{article}
\usepackage[preprint]{neurips_2026}
\workshoptitle{TS-LIMITS: Generalization for Time Series in Tight Settings: Latency, Inference, Memory, prIvacy and Sustainability}
\usepackage[utf8]{inputenc}
\usepackage[T1]{fontenc}
\usepackage{hyperref}
\usepackage{url}
\usepackage{booktabs}
\usepackage{microtype}
\usepackage{graphicx}
\usepackage{float}
\usepackage{tikz}
\usetikzlibrary{arrows.meta,positioning,fit}
\title{How Much Computation Does Every Forecast Need?\\Structured Block Skipping in Two Time-Series Foundation Models}

\author{%
  Max ~Bourgeat\\
  Polytechnique Montréal\\
  Montréal, QC, Canada \\
  \texttt{max.bourgeat@polymtl.ca}}

\begin{document}
\maketitle
\begin{abstract}
Time-series foundation models use the same network depth for every input, although forecasting windows differ in structure and difficulty. We test whether whole Transformer blocks can be removed at inference time in two pretrained models: Chronos-Bolt and TimesFM~2.5. Our bypasses avoid the original block computation and preserve the dense model exactly when all blocks are enabled. Masks are selected only on calibration windows and evaluated on 1,088 held-out windows from ETTh1, Traffic, Weather, and Exchange. At moderate sparsity, Chronos-Bolt with 7/11 candidate blocks and TimesFM with 16/20 blocks show little change in mean relative forecasting quality across datasets while reducing median A100 batch-1 latency by 11.2\% and 13.4\%. More aggressive TimesFM skipping accelerates inference further but substantially harms Traffic and Weather. Across both models, different equal-budget masks win on different windows, and a retrospective oracle significantly improves over the best static mask on every dataset. However, dataset-specific masks transfer inconsistently, and Chronos representations do not support reliable unseen-domain routing. These results establish cross-architecture structured redundancy and input-dependent opportunity, but not a deployable dynamic inference method.
\end{abstract}

\section{Introduction}
Pretrained time-series foundation models (TSFMs) have recently emerged as general-purpose models for zero- and few-shot forecasting across heterogeneous domains \citep{ansari2024chronos,das2024timesfm,rasul2024lagllama,woo2024moirai,goswami2024moment,ekambaram2024ttm}. Yet these models typically execute a fixed network for every input. This is undesirable when latency is constrained and potentially unnecessary when forecasting windows vary in predictability. Global compression reduces a network for all inputs. We ask a different question: does every input require the same structured subset of a pretrained TSFM?

We study executable whole-block skipping without fine-tuning. Our goal is not to propose a router, but to establish whether there is useful opportunity for one. We measure static quality-compute curves using calibration-selected masks and real GPU latency, compare equal-cardinality masks across datasets, and compute a retrospective per-window oracle. Using Chronos-Bolt, we then test whether temporal features or frozen representations can predict useful masks across domains.

We find: (1) removable computation is not model-specific: both Chronos-Bolt and the architecturally different decoder-only TimesFM admit moderate block skipping with real acceleration and limited aggregate quality change; (2) mask identity matters, especially in TimesFM, where the selected mask beats every one of 20 random masks at each sparse budget and dataset; and (3) both models have a significant per-window oracle advantage, but neither dataset identity nor tested Chronos features yield robust transferable routing. Static redundancy and input-dependent preference recur across models, while safe sparsity and routability remain architecture- and domain-dependent.

\section{Related work}
\paragraph{Time-series foundation models.}
Recent TSFMs pretrain on collections of heterogeneous time series to support transfer across datasets and domains. Chronos casts forecasting as token prediction \citep{ansari2024chronos}, TimesFM uses a decoder-only patched architecture \citep{das2024timesfm}, and Lag-Llama targets probabilistic zero-shot forecasting \citep{rasul2024lagllama}. Moirai and MOMENT broaden pretraining across datasets and tasks \citep{woo2024moirai,goswami2024moment}, while Tiny Time Mixers emphasize compact models and efficient transfer \citep{ekambaram2024ttm}. Our question is complementary: rather than train a smaller model, we ask how much of an already pretrained TSFM must execute for each forecast.

\paragraph{Structured and input-dependent computation.}
Reducing depth at inference has a substantial history outside time-series forecasting. LayerDrop trains Transformers so subnetworks of different depths can be extracted at inference \citep{fan2020layerdrop}, while DynaBERT learns width- and depth-adaptive subnetworks through training and distillation \citep{hou2020dynabert}. Input-dependent methods instead vary computation by example: adaptive computation time allocates a variable number of recurrent steps \citep{graves2016act}, SkipNet learns to bypass residual blocks \citep{wang2018skipnet}, and DeeBERT and FastBERT allow easier language inputs to exit Transformer stacks early \citep{xin2020deebert,liu2020fastbert}. These methods generally modify or train the model specifically for adaptive inference. We instead study frozen pretrained TSFMs without fine-tuning, directly bypass architecture-native blocks, measure realized GPU acceleration, and use a retrospective equal-budget oracle to test whether the identity of useful blocks varies across forecasting windows.

\section{Experimental design}
\paragraph{Models and executable units.} We use pretrained \texttt{amazon/chronos-bolt-base} and \texttt{google/timesfm-2.5-200m-transformers}. Chronos-Bolt is T5-style; its first encoder block remains mandatory and blocks 1--11 are candidates. TimesFM is decoder-only with 20 candidate Transformer blocks. A disabled candidate returns its input before attention and feed-forward computation. Counters verify that inactive blocks are not called. With all candidates active, controlled and untouched forecasts agree exactly. We do not remove weights, so all parameters remain resident.

\paragraph{Data and metrics.} We use ETTh1, Traffic, Weather, and Exchange \citep{zhou2021informer,wu2021autoformer}. Each example contains 512 context and 64 target steps. Selection uses disjoint calibration windows: 56 for ETTh1 and 64 for each other dataset. Evaluation uses 112, 512, 336, and 128 windows, respectively (1,088 total). We report normalized pinball loss over quantiles 0.1--0.9, with sMAPE and MASE as secondary metrics.

\paragraph{Selection and controls.} Marginal importance is the calibration-loss change after removing one block. Backward greedy selection repeatedly removes the active block giving the lowest remaining calibration loss. No evaluation window affects selection. Chronos uses 11, 9, 7, 4, and 2 active candidates; TimesFM uses 20, 16, 12, and 8. We evaluate 20 deterministic random masks at each sparse budget. Balanced-global masks minimize the equal-weight mean calibration loss across datasets. Dataset-specific masks are used only at the main oracle budgets: 7/11 for Chronos and 12/20 for TimesFM.

\paragraph{Oracle and routing diagnostics.} For each evaluation window, the oracle chooses the lowest-loss mask among four dataset-specific greedy masks and the balanced-global mask, all at equal cardinality. It is a retrospective upper bound, not an executable method. Confidence intervals use 2,000 bootstrap resamples of series/channel clusters. On Chronos only, small diagnostic classifiers use handcrafted temporal features or frozen context-only representations. Cross-validation groups windows from the same trajectory; leave-one-dataset-out (LODO) evaluation tests transfer to an unseen domain.

\paragraph{Systems measurement.} We benchmark one NVIDIA A100-SXM4-40GB using five warm-ups and 30 repetitions with CUDA synchronization. We report median batch-1 latency and batch-32 throughput. For Chronos, quality uses greedy masks whereas systems measurements use cardinality-matched marginal masks; because candidate blocks have identical shapes, Chronos timing measures active-block-count cost rather than mask-specific latency. TimesFM condition order is randomized. Resident parameter bytes and peak CUDA allocation are recorded separately.

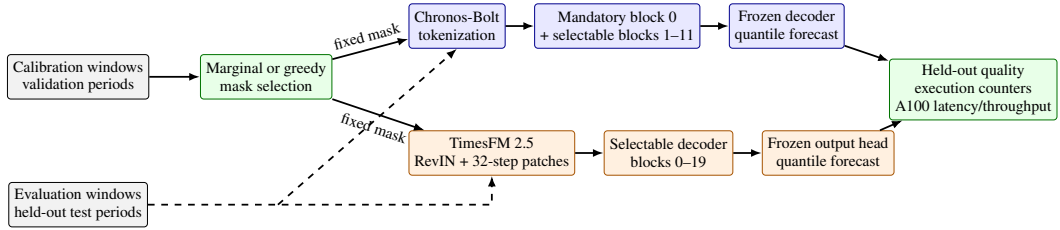
\begin{figure}[H]
\centering
\resizebox{\linewidth}{!}{%
\begin{tikzpicture}[
  font=\scriptsize,
  node distance=3.5mm and 4.5mm,
  box/.style={draw,rounded corners=1.5pt,align=center,minimum height=7mm,inner sep=2.5pt},
  data/.style={box,fill=gray!10},
  chronos/.style={box,fill=blue!9,draw=blue!55!black},
  timesfm/.style={box,fill=orange!12,draw=orange!65!black},
  shared/.style={box,fill=green!9,draw=green!45!black},
  arrow/.style={-{Latex[length=1.7mm]},thick},
  dashedarrow/.style={arrow,dashed}
]
\node[data] (cal) {Calibration windows\\validation periods};
\node[data,below=13mm of cal] (eval) {Evaluation windows\\held-out test periods};
\node[shared,right=8mm of cal] (select) {Marginal or greedy\\mask selection};

\node[right=18mm of eval] (void) {};

\node[chronos,right=12mm of select,yshift=8mm] (cpre) {Chronos-Bolt\\tokenization};
\node[chronos,right=of cpre] (cblocks) {Mandatory block 0\\+ selectable blocks 1--11};
\node[chronos,right=of cblocks] (chead) {Frozen decoder\\quantile forecast};

\node[timesfm,right=12mm of select,yshift=-12mm] (tpre) {TimesFM 2.5\\RevIN + 32-step patches};
\node[timesfm,right=of tpre] (tblocks) {Selectable decoder\\blocks 0--19};
\node[timesfm,right=of tblocks] (thead) {Frozen output head\\quantile forecast};

\node[shared,right=7mm of chead,yshift=-10mm] (measure) {Held-out quality\\execution counters\\A100 latency/throughput};

\draw[arrow] (cal) -- (select);
\draw[arrow] (select) -- node[above,sloped]{fixed mask} (cpre);
\draw[arrow] (select) -- node[below,sloped]{fixed mask} (tpre);
\draw[arrow] (cpre) -- (cblocks); \draw[arrow] (cblocks) -- (chead);
\draw[arrow] (tpre) -- (tblocks); \draw[arrow] (tblocks) -- (thead);
\draw[arrow] (chead) -- (measure); \draw[arrow] (thead) -- (measure);
\draw[dashedarrow] (void.east) -- (cpre.south);
\draw[dashedarrow] (eval.east) -| (tpre.south);

\end{tikzpicture}
}
\caption{Methodology for both frozen TSFMs. Calibration data select static masks without
test leakage; held-out windows then traverse genuinely bypassed architecture-native
blocks. The same forecasts support quality, execution, latency, throughput, and
retrospective equal-budget oracle analyses. Dashed arrows denote held-out inputs.}
\label{fig:method}
\end{figure}

Figure~\ref{fig:method} summarizes the shared protocol and the architecture-specific executable units.

\section{Static redundancy yields real acceleration}
Tables~\ref{tab:quality}--\ref{tab:systems} expose all tested static operating points. Chronos 7/11 preserves the aggregate metric, but its four dataset changes range from $-22.09\%$ to $+14.49\%$. TimesFM 16/20 is nearly neutral on average because its Weather improvement offsets degradation on Traffic. Thus limited aggregate change is not uniformly lossless. Interestingly, both models show the same dataset-level asymmetry at moderate sparsity: skipping degrades Traffic while improving Weather. This recurrence across two architectures reinforces that near-neutral aggregate quality can conceal substantial dataset-level variation. More aggressive TimesFM skipping crosses a sharp quality cliff, particularly on Traffic and Weather.

\begin{table}[t]\centering
\caption{Per-dataset static quality change relative to dense (normalized pinball loss; lower is better). Mean is the unweighted mean of the four within-dataset relative changes.}\label{tab:quality}\small
\begin{tabular}{llrrrrr}\toprule
Model & Active & ETTh1 & Traffic & Weather & Exchange & Mean \\\midrule
Chronos-Bolt & 11/11 & 0.00\% & 0.00\% & 0.00\% & 0.00\% & 0.00\% \\
 & 9/11 & $+1.75\%$ & $+1.61\%$ & $-4.96\%$ & $-0.21\%$ & $-0.45\%$ \\
 & 7/11 & $+2.73\%$ & $+14.49\%$ & $-22.09\%$ & $+0.19\%$ & $-1.17\%$ \\
 & 4/11 & $+19.55\%$ & $+41.43\%$ & $+47.13\%$ & $+10.96\%$ & $+29.77\%$ \\
 & 2/11 & $+18.96\%$ & $+70.81\%$ & $+48.99\%$ & $+1.01\%$ & $+34.94\%$ \\\midrule
TimesFM 2.5 & 20/20 & 0.00\% & 0.00\% & 0.00\% & 0.00\% & 0.00\% \\
 & 16/20 & $-0.02\%$ & $+8.15\%$ & $-8.29\%$ & $+0.11\%$ & $-0.01\%$ \\
 & 12/20 & $+4.42\%$ & $+49.72\%$ & $+32.54\%$ & $+6.81\%$ & $+23.37\%$ \\
 & 8/20 & $+4.81\%$ & $+65.84\%$ & $+85.91\%$ & $+7.62\%$ & $+41.05\%$ \\\bottomrule
\end{tabular}\end{table}

\begin{table}[t]\centering
\caption{Measured A100 systems results. Parentheses give change from the model's dense baseline; throughput uses batch 32.}\label{tab:systems}\small
\begin{tabular}{llrr}\toprule
Model & Active & Batch-1 latency & Batch-32 throughput \\\midrule
Chronos-Bolt & 11/11 & 32.32 ms (0.0\%) & 944.6 series/s (0.0\%) \\
 & 9/11 & 30.53 ms ($-5.5\%$) & 1001.7 series/s ($+6.0\%$) \\
 & 7/11 & 28.71 ms ($-11.2\%$) & 1065.4 series/s ($+12.8\%$) \\
 & 4/11 & 25.79 ms ($-20.2\%$) & 1177.4 series/s ($+24.6\%$) \\
 & 2/11 & 23.91 ms ($-26.0\%$) & 1263.5 series/s ($+33.8\%$) \\\midrule
TimesFM 2.5 & 20/20 & 61.68 ms (0.0\%) & 455.1 series/s (0.0\%) \\
 & 16/20 & 53.40 ms ($-13.4\%$) & 531.2 series/s ($+16.7\%$) \\
 & 12/20 & 45.22 ms ($-26.7\%$) & 633.3 series/s ($+39.2\%$) \\
 & 8/20 & 37.03 ms ($-40.0\%$) & 774.1 series/s ($+70.1\%$) \\\bottomrule
\end{tabular}\end{table}

Greedy is consistently better than marginal ranking on the TimesFM aggregate curve and beats the best of 20 equal-cardinality random masks in all 12 budget--dataset comparisons. Chronos selection is less uniformly separated and does not beat every sampled mask. Thus mask choice matters, but greedy is not proven optimal.

\begin{figure}[t]\centering
\includegraphics[width=.68\linewidth]{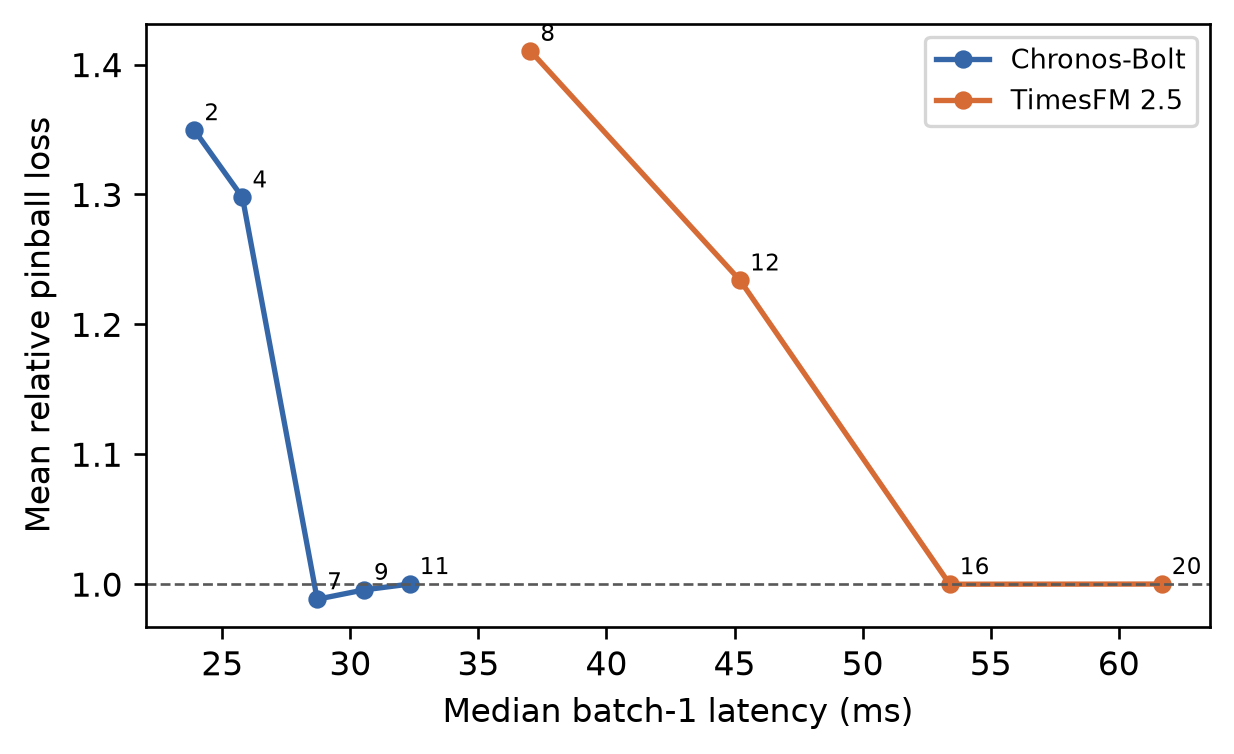}
\caption{Quality-latency trade-off. Loss is normalized within model and dataset before averaging; latency is synchronized A100 batch-1 inference.}\label{fig:pareto}
\end{figure}

Figure~\ref{fig:pareto} confirms that removed blocks translate to measured acceleration. These are compute gains, not resident-model-memory gains: execution counters show bypassed blocks are not called, but all weights remain loaded. These are compute rather than model-memory gains: skipped blocks are not executed, but their weights remain resident. Peak CUDA allocation is unchanged across budgets within each batch regime.

\section{Mask preference varies across windows}
Dataset-specific masks do not define stable specializations. TimesFM's five 12/20 masks have pairwise Jaccard similarity 0.60--0.85, yet a dataset's own mask is not reliably best: the Traffic mask is best on Exchange, the ETTh1 mask is best on Weather, and the Weather mask worsens Weather by 70.6\%. Chronos shows the same qualitative instability. Dataset identity is therefore an inadequate routing rule.

\begin{figure}[t]\centering
\includegraphics[width=.92\linewidth]{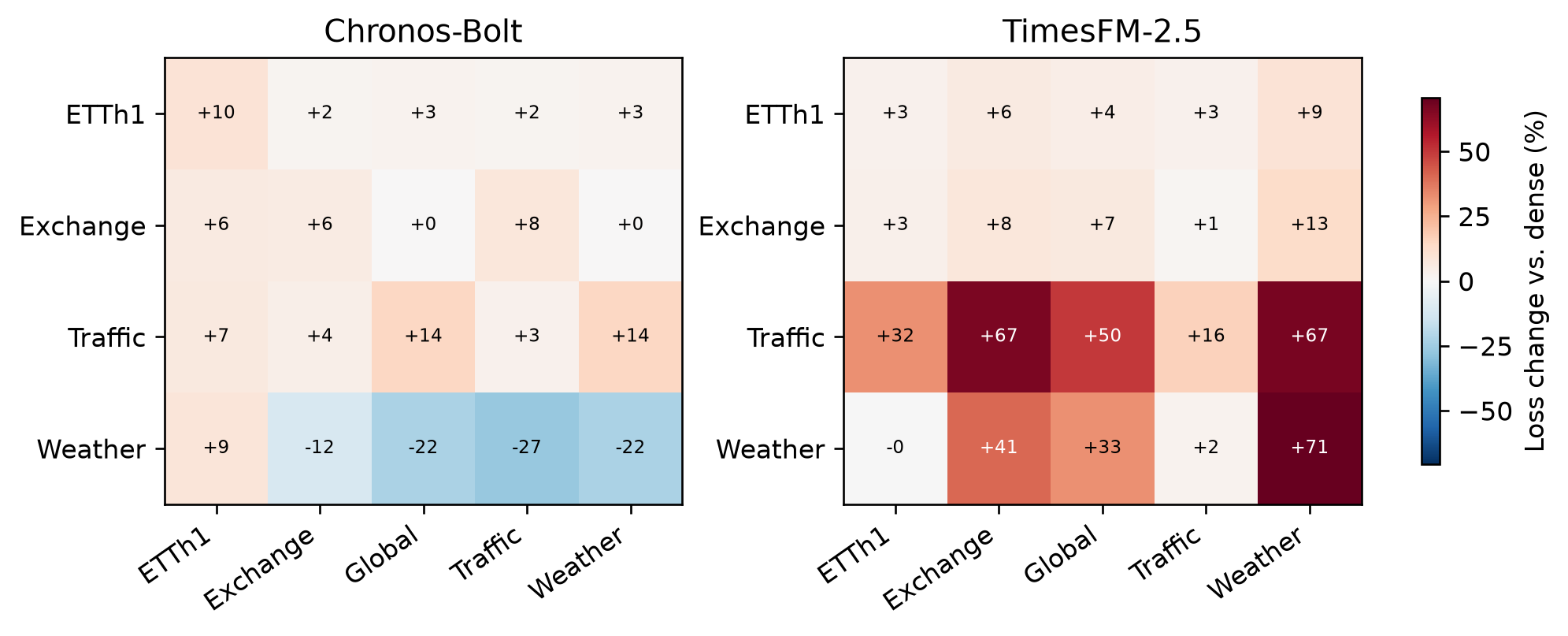}
\caption{Cross-dataset transfer at the main budget (7/11 Chronos, 12/20 TimesFM).
Each cell is pinball-loss change relative to dense on the evaluation dataset; lower is
better. Columns denote mask-selection origin.}\label{fig:transfer}
\end{figure}

Figure~\ref{fig:transfer} makes this instability visible after normalizing every cell
to its dataset's dense loss. A dataset-specialized pattern would place the best value
on the native diagonal. Instead, off-diagonal masks frequently match or beat the native
mask, and degradation is asymmetric. This does not imply one universal sparse network;
it shows that dataset labels alone do not identify a consistently preferred one.

Despite weak specialization, the per-window oracle improves significantly over the best static equal-budget mask on every dataset and both models (Table~\ref{tab:oracle}). Relative gains range from 7.5--21.7\% for Chronos and 3.2--17.1\% for TimesFM; all cluster-bootstrap intervals exclude zero. All five masks win on every TimesFM dataset. Effective winning-mask counts are 4.68 on ETTh1, 4.68 on Weather, and 4.95 on Exchange. Traffic is less dynamic: its native mask wins 79.1\% of windows and oracle gain is 3.2\%. Input dependence is cross-model but heterogeneous.

\begin{figure}[t]\centering
\includegraphics[width=.96\linewidth]{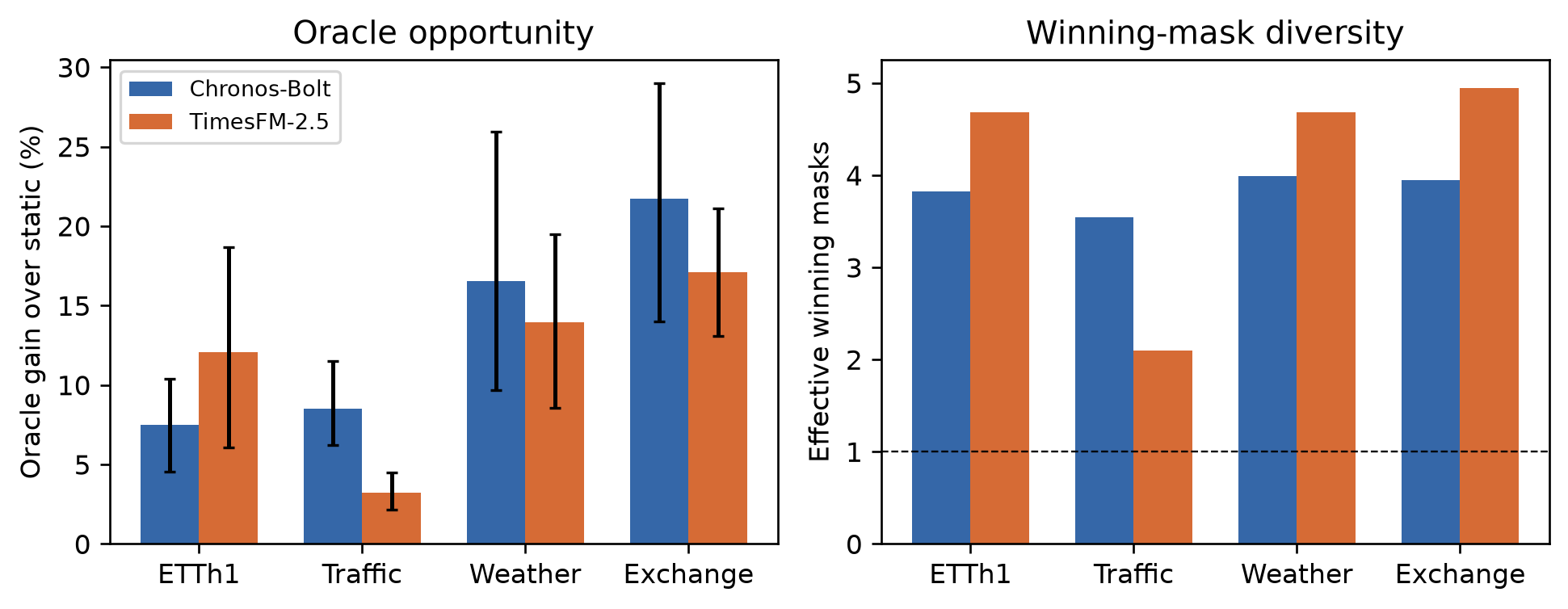}
\caption{Oracle opportunity and diversity. Left: relative gain over the best static
equal-budget mask with 95\% series-cluster bootstrap intervals. Right: exponential
entropy of the winning-mask distribution (maximum five); one denotes a single winner.}
\label{fig:oracle}\end{figure}

Figure~\ref{fig:oracle} separates oracle magnitude from diversity. A positive gain with
one effective winner would favor static selection; instead, most dataset-model pairs
combine positive gain with several winners. TimesFM Traffic is the clearest exception.
For Chronos, tied wins are divided evenly among tied masks before entropy is converted
to an effective count, avoiding treatment of a tie set as a new mask.

\begin{table}[t]\centering
\caption{Oracle gain over the evaluation-best static equal-budget mask. All absolute-gain 95\% series-cluster bootstrap intervals exclude zero.}\label{tab:oracle}\small
\begin{tabular}{llrrr}\toprule
Model & Dataset & Static & Oracle & Relative gain \\\midrule
Chronos-Bolt & ETTh1 & .2076 & .1921 & 7.5\% \\
 & Traffic & .1596 & .1460 & 8.5\% \\
 & Weather & .2056 & .1717 & 16.5\% \\
 & Exchange & .0158 & .0123 & 21.7\% \\\midrule
TimesFM 2.5 & ETTh1 & .2178 & .1915 & 12.1\% \\
 & Traffic & .1763 & .1706 & 3.2\% \\
 & Weather & .1745 & .1502 & 13.9\% \\
 & Exchange & .0152 & .0126 & 17.1\% \\\bottomrule
\end{tabular}\end{table}

Chronos diagnostics test whether the oracle can be approached cheaply. Handcrafted features are harmful for mask selection. Frozen representations contain more within-domain signal: block-5 mean-plus-standard-deviation features reach 45.1\% balanced mask accuracy and recover 29.5\% of static-to-oracle gain, with a cluster bootstrap interval of 8.2--46.3\%. However, LODO transfer mostly approaches chance and usually worsens selector loss. Later representations also consume part of the computation one hopes to save. The oracle demonstrates opportunity, not a practical router.

\section{Discussion and limitations}
Executable redundancy, real acceleration, and per-window preference occur in both a T5-style Chronos model and decoder-only TimesFM. Their magnitudes differ: TimesFM quality degrades sharply below 80\% active on Traffic and Weather, while Chronos tolerates 7/11 better on average. Static sparsity must be calibrated per architecture and domain.

Our scope is limited to two models, four univariate benchmarks, and one GPU type. Greedy masks are selected candidates, not optimal subnetworks. The oracle uses held-out losses retrospectively, and failed LODO diagnostics show that opportunity does not imply routability. Block skipping leaves weights resident, so we establish latency and throughput savings, not model-memory or energy savings. We do not report an edge-device benchmark. A dynamic system would require a cheap, transferable signal and end-to-end measurement including routing overhead; our experiments do not satisfy that requirement.

\section{Conclusion}
Two architecturally different TSFMs perform removable block computation, and different equal-budget subnetworks are useful for different windows. Moderate static skipping reduces A100 latency by 11--13\% with little aggregate quality change, despite substantial dataset-level variation. A retrospective oracle improves over the best static mask in both models, but dataset-specific transfer and unseen-domain routing remain unreliable. We find structured redundancy and input-dependent opportunity, not a deployable conditional-computation system.

\bibliographystyle{plainnat}
\bibliography{references-expanded}

@article{ansari2024chronos,
 title={Chronos: Learning the Language of Time Series},
 author={Ansari, Abdul Fatir and Stella, Lorenzo and Turkmen, Caner and Zhang, Xiyuan and Mercado, Pedro and Shen, Huibin and Shchur, Oleksandr and Rangapuram, Syama Syndar and Pineda Arango, Sebastian and Kapoor, Shubham and Zschiegner, Jasper and Maddix, Danielle C. and Mahoney, Michael W. and Torkkola, Kari and Wilson, Andrew Gordon and Bohlke-Schneider, Michael and Wang, Yuyang},
 journal={Transactions on Machine Learning Research}, year={2024}}

@inproceedings{das2024timesfm,
 title={A Decoder-Only Foundation Model for Time-Series Forecasting},
 author={Das, Abhimanyu and Kong, Weihao and Sen, Rajat and Zhou, Yichen},
 booktitle={International Conference on Machine Learning}, year={2024}}

@inproceedings{zhou2021informer,
 title={Informer: Beyond Efficient Transformer for Long Sequence Time-Series Forecasting},
 author={Zhou, Haoyi and Zhang, Shanghang and Peng, Jieqi and Zhang, Shuai and Li, Jianxin and Xiong, Hui and Zhang, Wancai},
 booktitle={AAAI Conference on Artificial Intelligence}, year={2021}}

@inproceedings{wu2021autoformer,
 title={Autoformer: Decomposition Transformers with Auto-Correlation for Long-Term Series Forecasting},
 author={Wu, Haixu and Xu, Jiehui and Wang, Jianmin and Long, Mingsheng},
 booktitle={Advances in Neural Information Processing Systems}, year={2021}}

@article{rasul2024lagllama,
 title={Lag-Llama: Towards Foundation Models for Probabilistic Time Series Forecasting},
 author={Rasul, Kashif and Ashok, Arjun and Williams, Andrew Robert and Ghonia, Hena and Bhagwatkar, Rishika and Khorasani, Arian and Bayazi, Mohammad Javad Darvishi and Adamopoulos, George and Riachi, Roland and Hassen, Nadhir and Bilo\v{s}, Marin and Garg, Sahil and Schneider, Anderson and Chapados, Nicolas and Drouin, Alexandre and Zantedeschi, Valentina and Nevmyvaka, Yuriy and Rish, Irina},
 journal={arXiv preprint arXiv:2310.08278}, year={2024}}

@InProceedings{woo2024moirai,
  title = 	 {Unified Training of Universal Time Series Forecasting Transformers},
  author =       {Woo, Gerald and Liu, Chenghao and Kumar, Akshat and Xiong, Caiming and Savarese, Silvio and Sahoo, Doyen},
  booktitle = 	 {Proceedings of the 41st International Conference on Machine Learning},
  pages = 	 {53140--53164},
  year = 	 {2024},
  editor = 	 {Salakhutdinov, Ruslan and Kolter, Zico and Heller, Katherine and Weller, Adrian and Oliver, Nuria and Scarlett, Jonathan and Berkenkamp, Felix},
  volume = 	 {235},
  series = 	 {Proceedings of Machine Learning Research},
  month = 	 {21--27 Jul},
  publisher =    {PMLR},
  url = 	 {https://proceedings.mlr.press/v235/woo24a.html}
}

@inproceedings{goswami2024moment,
 title={MOMENT: A Family of Open Time-series Foundation Models},
 author={Goswami, Mononito and Szafer, Konrad and Choudhry, Arjun and Cai, Yifu and Li, Shuo and Dubrawski, Artur},
 booktitle={International Conference on Machine Learning},
 pages={16115--16152}, year={2024}}

@article{ekambaram2024ttm,
  title={Tiny time mixers (ttms): Fast pre-trained models for enhanced zero/few-shot forecasting of multivariate time series},
  author={Ekambaram, Vijay and Jati, Arindam and Dayama, Pankaj and Mukherjee, Sumanta and Nguyen, Nam H and Gifford, Wesley M and Reddy, Chandra and Kalagnanam, Jayant},
  journal={Advances in Neural Information Processing Systems},
  volume={37},
  pages={74147--74181},
  year={2024}
}

@inproceedings{fan2020layerdrop,
 title={Reducing Transformer Depth on Demand with Structured Dropout},
 author={Fan, Angela and Grave, Edouard and Joulin, Armand},
 booktitle={International Conference on Learning Representations}, year={2020}}

@inproceedings{hou2020dynabert,
 title={DynaBERT: Dynamic BERT with Adaptive Width and Depth},
 author={Hou, Lu and Huang, Zhiqi and Shang, Lifeng and Jiang, Xin and Chen, Xiao and Liu, Qun},
 booktitle={Advances in Neural Information Processing Systems}, year={2020}}

@article{graves2016act,
 title={Adaptive Computation Time for Recurrent Neural Networks},
 author={Graves, Alex},
 journal={arXiv preprint arXiv:1603.08983}, year={2016}}

@inproceedings{wang2018skipnet,
 title={SkipNet: Learning Dynamic Routing in Convolutional Networks},
 author={Wang, Xin and Yu, Fisher and Dou, Zi-Yi and Darrell, Trevor and Gonzalez, Joseph E.},
 booktitle={European Conference on Computer Vision}, pages={409--424}, year={2018}}

@inproceedings{xin2020deebert,
 title={DeeBERT: Dynamic Early Exiting for Accelerating BERT Inference},
 author={Xin, Ji and Tang, Raphael and Lee, Jaejun and Yu, Yaoliang and Lin, Jimmy},
 booktitle={Proceedings of the 58th Annual Meeting of the Association for Computational Linguistics},
 pages={2246--2251}, year={2020}}

@inproceedings{liu2020fastbert,
 title={FastBERT: a Self-distilling BERT with Adaptive Inference Time},
 author={Liu, Weijie and Zhou, Peng and Wang, Zhiruo and Zhao, Zhe and Deng, Haotang and Ju, Qi},
 booktitle={Proceedings of the 58th Annual Meeting of the Association for Computational Linguistics},
 pages={6035--6044}, year={2020}}
\end{document}